\documentclass[letterpaper, 10 pt, conference]{ieeeconf}  
\usepackage{amsmath, amssymb}
\usepackage{booktabs}
\usepackage{multirow}
\usepackage{xspace}
\usepackage{sidecap}
\usepackage{wrapfig}
\usepackage{hyperref}
\usepackage{cite}
\usepackage{graphicx}
\usepackage{xcolor}

\IEEEoverridecommandlockouts                             
\title{\LARGE \bf
Exploiting Local Geometry for Feature and Graph Construction 
for Better 3D Point Cloud Processing with Graph Neural Networks
}

\author{Siddharth Srivastava$^{1}$ and Gaurav Sharma$^{2}$

\thanks{$^{1}$Siddharth Srivastava is with Centre for Development of Advanced Computing, Noida, India}%
\thanks{$^{2}$Gaurav Sharma is with TensorTour and IIT Kanpur. A part of this work was done when Gaurav Sharma was with NEC Labs America.}%
}
\def\onedot{.\xspace}
\def\wrt{w.r.t.\xspace}

\begin{document}

\maketitle
\thispagestyle{empty}
\pagestyle{empty}

\def\etal{et al\onedot}
\def\etc{etc\onedot}
\def\ie{i.e\onedot}
\def\eg{e.g\onedot}
\def\cf{cf\onedot}
\def\vs{vs\onedot}
\def\knn{\ensuremath{k}-NN\xspace}
\def\L{\ensuremath{\mathcal{L}}\xspace}
\def\T{\ensuremath{\mathcal{T}}\xspace}
\def\x{\ensuremath{\mathbf{x}}\xspace}
\def\p{\ensuremath{\mathbf{p}}\xspace}
\def\y{\ensuremath{\mathbf{y}}\xspace}
\def\X{\ensuremath{\mathbf{X}}\xspace}
\def\D{\ensuremath{D}\xspace}
\def\R{\ensuremath{\mathbb{R}}\xspace}
\def\G{\ensuremath{\mathcal{G}}\xspace}
\def\E{\ensuremath{\mathcal{E}}\xspace}
\def\V{\ensuremath{\mathcal{V}}\xspace}

\newcommand{\heading}[1]
{
\ \vspace{-.8em} \\
\textbf{#1}
}

\begin{abstract}
We propose simple yet effective improvements in point representations and local neighborhood
graph construction within the general framework of graph neural networks (GNNs) for 3D point cloud
processing. As a first contribution, we propose to augment the vertex representations with important
local geometric information of the points, followed by nonlinear projection using a MLP. As a second
contribution, we propose to improve the graph construction for GNNs for 3D point clouds. The
existing methods work with a \knn based approach for constructing the local neighborhood graph. We
argue that it might lead to reduction in coverage in case of dense sampling by sensors in some
regions of the scene. The proposed methods aims to counter such problems and improve coverage in
such cases. As the traditional GNNs were designed to work with general graphs, where vertices may have no geometric interpretations, we see both our proposals as
augmenting the general graphs to incorporate the geometric nature of 3D point clouds. While being
simple, we demonstrate with multiple challenging benchmarks, with relatively clean
CAD models, as well as with real world noisy scans, that the proposed method achieves
state of the art results on benchmarks for 3D classification (ModelNet40) , part segmentation (ShapeNet) and semantic segmentation (Stanford 3D Indoor Scenes Dataset). We also show that the proposed network achieves faster training convergence, \ie $\sim40\%$
less epochs for classification. The project details are available at \url{https://siddharthsrivastava.github.io/publication/geomgcnn/}
\end{abstract}

\vspace{-0.5ex}
\section{Introduction}
\label{secIntro}
\vspace{-0.5ex}

Significant progress has been made recently towards designing neural network architectures for
directly processing unstructured 3D point clouds~\cite{ioannidou2017deep}. Since 3D point clouds are
obtained as the native output from commonly available scanning sensors, the ability to process them and
apply machine learning methods directly to them is interesting, and allows for many useful
applications~\cite{liu2019deep}. Like many other input modalities, both handcrafted and learning
based methods are available for extracting meaningful information from 3D point clouds. The current
state-of-the-art techniques for extracting such information are mostly based on deep neural
networks. 

\begin{figure}
\centering
\includegraphics[width=0.9\columnwidth]{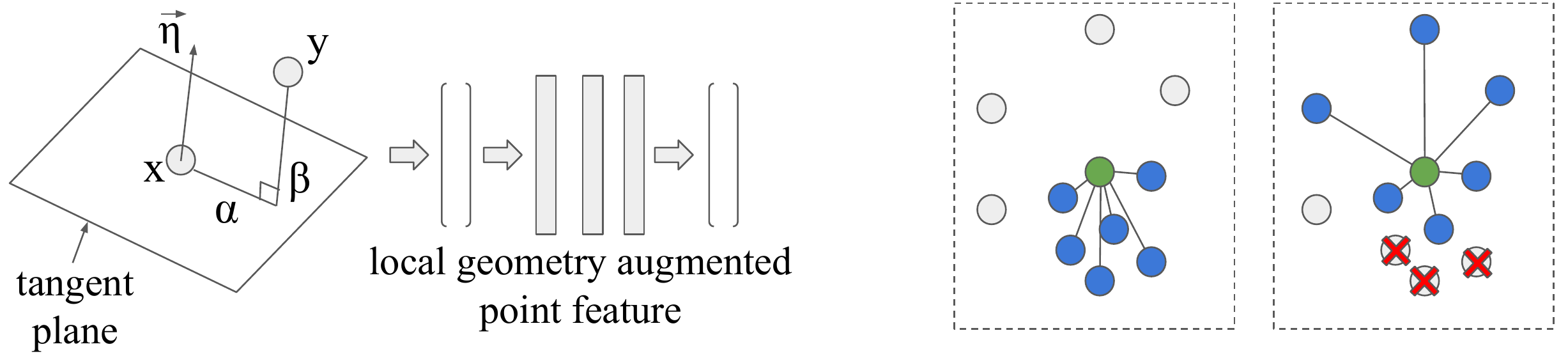}
\caption{\small{We propose two components which work within the framework of graph neural networks for 3D
point cloud processing. First, we propose to augment the point representations, not just by point
features (\eg coordinates) but also with local geometric information such as the in and out plane
distances $\alpha, \beta$ (shown in left), \wrt the tangent plane of point $\x$, to the nearby
points, \eg $\y$.  We further propose to nonlinearly project the resulting feature with a small
neural network. Second, we propose to construct the local neighborhood graph in a geometrically
aware way.  While in the usual \knn based construction, the local graph can get biased towards a
certain local neighborhood in the case of selective dense sampling, the proposed
method aims to avoid that (shown in the two right blocks), by selecting the neighbors in a
geometrically aware fashion. Best viewed in color.}}
\label{figIllus}
\end{figure}

As the point clouds are unordered, it is difficult to use CNNs with them. The popular ways of
overcoming the problem of unstructured point clouds have been to either (i) voxelize them, \eg~\cite{maturana2015voxnet} or (ii) learn a permutation invariant mapping, \eg~\cite{qi2016pointnet}. Point clouds can also be seen as graphs with potential edges defined between
neighboring points. Therefore, utilizing graph neural networks (GNN) for processing point cloud data
emerges as a natural choice. Most of the earlier GNN methods~\cite{wu2019comprehensive} considered
the points in isolation, and did not explicitly utilize the information from local neighborhoods.
While such local geometric information can be helpful for higher level tasks, these methods expected
the network to learn the important intricacies directly from isolated points as inputs. However, the Dynamic EdgeConv network~\cite{wang2019dynamic} showed that local geometric properties are important for learning feature representations from point clouds. They proposed
to use the differences between the 3D points and their neighbors, which can be interpreted as a 3D
direction vector, to encode the local structure. In addition, they converted the point cloud to a
graph using a \knn based construction, which is another way of encoding local geometric structure
for use by the method. By doing this they were able to close the gap between GNN based methods and
other state-of-the-art point cloud processing methods.

Along the lines of Dynamic EdgeConv, we believe that presenting more geometric information encoded
in a task relevant way is important for the success of GNNs for point cloud processing. The generic
graph networks \cite{wu2019comprehensive} were designed to work with general graphs, \eg citation
networks, social network graphs \etc In such graphs there is no notion of geometric relations
between vertices. However, point cloud data is inherently geometric and one could expect the
performance to increase when the two natural properties of the point cloud data, \ie graph like
structure, and geometric information, are co-exploited. We see Dynamic EdgeConv work as a step
along, and propose to go further in this direction.

We propose two novel, simple but extremely effective, components within the framework of general
GNNs based on the intuitions above as shown in Figure \ref{figIllus}. First, we propose to encode detailed local geometric
information, inspired by traditional handcrafted 3D descriptors, about the points. Further, to
convert the raw geometric information to a form which is easily exploited, we propose to use a multi
layer perceptron network (MLP) to nonlinearly project the point features into another space. We then
propose to present the projected representations as inputs to the graph network.  This goes beyond
the usual recommendation in previous works, where the possibility of adding more information about
the point, \eg color, depth, in addition to its 3D coordinates, has been often mentioned. We encode
not only more information about the point, but also encode the information derived from the local
neighborhood, around the point, as shown beneficial by previous works in handcrafted 3D descriptors.
We also project such information nonlinearly to make it better exploitable by the GNN. While, in
theory such information can be extracted by the neural network from 3D point coordinate inputs, we
believe that presenting this information upfront allows the network to start from a better point,
and even regularizes the learning. 

Second, we propose a geometrically informed graph construction which goes beyond the usual \knn
based construction. We constrain the sequential selection of points to be added to a local graph
based on their angles and distances with already selected points, while considering the point of
interest as the center of reference. This increases the coverage of the local connectivity and
especially addresses cases where the points might be very densely sampled in one neighborhood, while
relatively sparsely sampled in another.

We show empirically that both the contributions, while being relatively simple, are significant and
help improve the performance of the baseline GNN. We report results on
multiple publicly available challenging benchmarks of point clouds including both CAD models and real world 3D scans.

The results on the real world datasets help us make
the point that the local geometry extraction methods work despite the noise in real world scans.

In summary, the contributions of this work are:
\begin{itemize}
\setlength\itemsep{-.1em}
    \item We propose to represent the vertices in a point cloud derived graph, corresponding to the
    3D points, with geometric properties computed with respect to their local neighborhoods. 
    \item We propose a geometrically informed construction of the local neighborhood graph, instead
    of using a na\"ive \knn based initialization.  
    \item We give extensive empirical results on five public
    benchmarks \ie classification on ModelNet40 \cite{wu20153d}, part segmentation on ShapeNet \cite{yi2016scalable} and semantic segmentation on ScanNet \cite{dai2017scannet}, S3DIS \cite{armeni20163d} and Paris-Lille-3D \cite{roynard2018paris}, reporting state-of-the-art results on three of them,  while competing with both graph based as well as other neural networks for point cloud processing.  We also give exhaustive ablation results and indicative qualitative results to fully validate the method.
\end{itemize}

\section{Related Works} \label{sec:relatedworks}
\vspace{-0.5ex}
Many tasks in 3D, such as classification and segmentation, can be
solved by exploiting the local geometric structure of point clouds. In earlier efforts, handcrafted
descriptors did this by encoding various properties of local regions in point clouds.  The
handcrafted descriptors, in general, fall into two categories. First are based on spatial
distribution of the points, while the second directly utilize geometric properties of the points on
surface. 
 
Most of the handcrafted descriptors are based on finding a local reference frame which provides
robustness to various types of transformations of the point cloud. Spin Image
\cite{johnson1998surface} uses in-plane and out-plane distance of neighborhood points within the
support region of the keypoint to discretize the space and construct the descriptor.
Local Surface Patches \cite{chen20073d, chen2007human} utilizes shape index
\cite{koenderink1992surface} and angle between normals to accumulate points forming the final
descriptor. 

3D Binary Signatures \cite{srivastava20163d} uses binary
comparisons of geometrical properties in a local neighborhood. We refer the reader to Guo \etal
\cite{guo2016comprehensive} for a comprehensive survey of handcrafted 3D local descriptors.

Further, learning based methods have also been applied to obtain both local and global
descriptors in 3D. 
Compact Geometric Features (CGF) \cite{khoury2017learning} learns an
embedding to a low dimensional space where the similar points are closer. 
DLAN \cite{furuya2016deep}
performs aggregation of local 3D features using deep neural network for the task of retrieval. 
DeepPoint3D \cite{srivastava2019deeppoint3d} learns discriminative local descriptors by using deep metric learning. 

While 3D local descriptors are still preferred for tasks such as point cloud registration, 3D deep
networks trained for generating a global descriptor have provided state-of-the-art results on
various classification, segmentation and retrieval benchmarks. The deep networks applied to these
tasks are primarily based on volumetric representations (voxels, meshes, 
point clouds) and/or multi-view renderings of 3D shapes. We focus on techniques which directly
process raw 3D point clouds as we work with them as well. 

PointNet \cite{qi2016pointnet} directly processes 3D point clouds by aggregating information using a
symmetric function. 
PointNet++ \cite{qi2017pointnet++}
applies hierarchical learning to PointNet to obtain geometrically more robust features. 
Kd-Net 
\cite{klokov2017escape} constructs a network of kd-trees for parameter
learning and sharing. PCNN~\cite{atzmon2018point}, RS-CNN~\cite{liu2019relation}, LP-3DCNN~\cite{kumawat2019lp}, MinkowskiNet~\cite{choy20194d}, FKAConv~\cite{boulch2020lightconvpoint}, PointASNL~\cite{yan2020pointasnl} propose
modifications to convolution kernel or layers to produce features.  

Recently, graph based neural networks have shown good performance on such unstructured data~\cite{li2019deepgcns, wang2019graph, liang2019hierarchical}.
Dynamic Edge Convolutional Neural Network (DGCNN) \cite{wang2019dynamic} showed that
by representing the local neighborhood of a point as a graph, it can achieve better performance than the previous techniques.    Moreover, many previous works such as PointNet, PointNet++ become a special case with various Edge Convolution functions. LDGCNN \cite{zhang2019linked} extends DGCNN by hierarchically linking features from the dynamic
graphs in DGCNN. Authors in ~\cite{lei2020spherical,lei2020seggcn} propose modifications to kernels to capture geoemtric relationships between points. Our proposals are mainly within the purview of GNNs, however we extensively
compare with most of the existing 3D point cloud based methods and show better results.

\begin{figure*}[h]
\centering
\includegraphics[width=.8\textwidth]{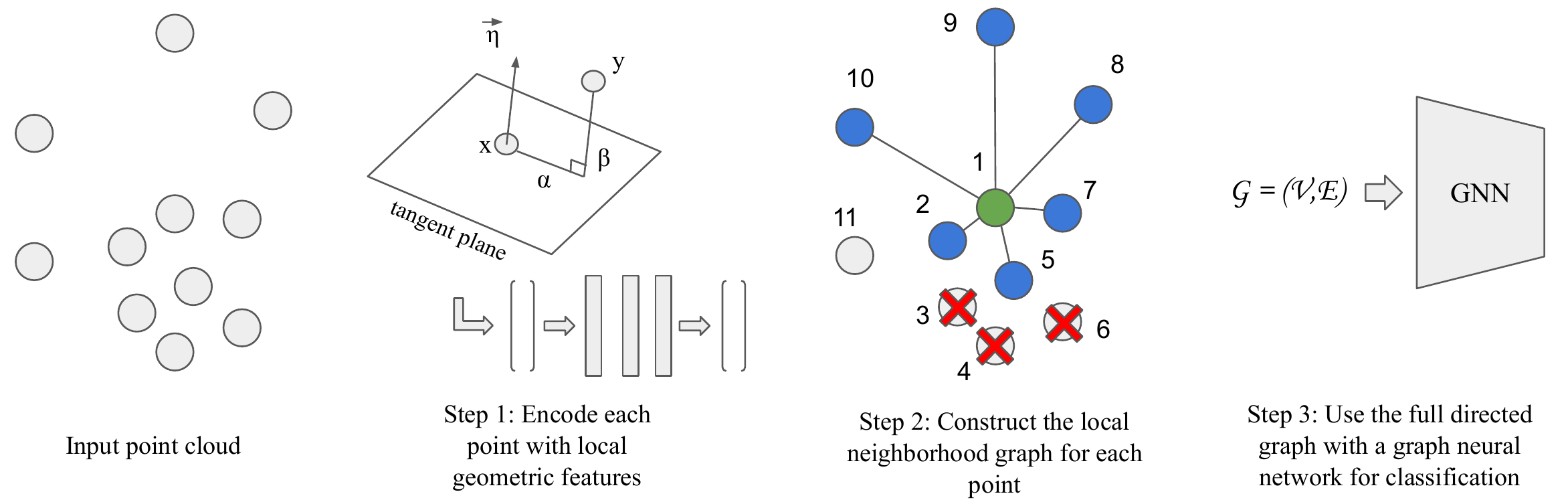}
\vskip -0.1in
\caption{\small{Illustration of of the proposed method. The input point cloud is processed point by point
to generate an eventual graph, with vertices represented as real vectors and having directed edges
denoting local neighborhoods. The representation includes local geometric properties of the points
(which is relative to other points in their neighborhood), along with their coordinates. The graph
formulation is geometrically aware and counters oversampling in some regions of the space. The
finally constructed graph can then be used with a graph neural network for supervised learning, 
\eg for classification and part segmentation. Best viewed in color.}}
\label{figBlockDiag}
\end{figure*}

\vspace{-1ex}
\section{Approach}
\label{sec:approach}

We are interested in processing a set \X, \ie a point cloud, of 3D points. We follow the general
graph neural network (GNN) based processing, where we work with a directed graph $\G=(\V,\E)$ with
the points being the vertices \V and the edges being potential connections between them. The
vertices in \V are represented by a \D dimensional feature vector, which is often just the 3D
coordinate of the points, and the set of edges is constructed usually with a \knn based graph
construction algorithm. We now describe our two contributions pertaining to (i) the representation
of the points using local geometric information and (ii) constructing the graph edges \E with a
geometrically constrained algorithm. Figure~\ref{figBlockDiag} gives an overall block diagram of our
full system.

\subsection{Local Geometric Representation of Points} \label{subsec:localrep}

Majority of the GNN based previous works use a basic representations of points, \ie their 3D
coordinates, and feed that to the GNN to extract more complex interactions between neighboring
points. Since GNNs, \eg \cite{wu2019comprehensive}, were designed for general graphs which may not
have any geometric properties, \eg social network graphs, they do not have any explicit mechanism to
encode geometry. More recent works on point cloud processing have started exploiting the geometric
nature of 3D data more explicitly. As a specific case, EdgeConv \cite{wang2019dynamic} uses an edge
function of the form  

$
h_{\boldsymbol{\Theta}}(\x_i, \x_j) = \bar h_{\boldsymbol{\Theta}}(\x_i, \x_j - \x_i).
$

where $\x_i$ is the point under consideration, and $\x_j$ is one of its neighbors. $h_{\boldsymbol{\Theta}}$ is a nonlinear function and $\boldsymbol{\Theta}$ is a set of learnable parameters.
This function thus specifically uses a global description of the point, \ie the point coordinates
themselves, and a local description, which is the direction vector from the point to a nearby point.
Providing such explicit local geometric description as input to the network layers helps EdgeConv
outperform the baseline GNNs.

We go further in this direction and propose to use detailed geometric description of the local
neighborhood of the point. Significant ingenuity was spent on the design of handcrafted 3D features
(see \cite{guo2016comprehensive} for a comparison), before deep neural networks became competitive
for 3D point cloud processing. We take inspiration from such works in deriving the geometric
descriptions of the points.

We include the following components in the representation of each point. As used in previous works,
the first part is plain 3D coordinates of the point $(x,y,z)$, followed by the normal vector
$\vec{\eta} = (\eta_x, \eta_y, \eta_z)$ of the point.
Then we take inspiration from the Spin Image descriptor \cite{johnson1998surface} and add the in-plane and
out-plane distances $\alpha, \beta$ \wrt the point. 
Finally, inspired by the Local Surface Patch Descriptor \cite{chen20073d,chen2007human}, we include
the shape index ($\gamma$) as well.

Hence, compared to a simple 3D coordinate based description of the points, we represent them with a
9D descriptor $\p = (x,y,z, \eta_x,\eta_y,\eta_z, \alpha,\beta, \gamma)$ containing more local contextual
geometric information. We then further pass this 9D descriptor through a small 
neural network $\phi(\cdot)$ to nonlinearly project it into another space which is more adapted to
be used with the GNN. Finally, we pass the nonlinearly projected point representation $\phi(\p)$ to
the graph neural network. 

While one can argue that such augmentation may be redundant as the GNN should be able to extract it
out from 3D coordinates only, we postulate that providing these, successful handcrafted descriptor
inspired and nonlinearly projected features, makes the job of the GNN easier. Another argument could
be raised that these descriptors might be very task specific and hence might not be needed for a
task of interest. In that case, we would argue that it should be possible for GNN to essentially
ignore them by learning zero weights for weights/edges corresponding to these input units in the
neural network. While this is a seemingly simple augmentation, we show empirically in
Section~\ref{secExp} that, in fact, adding such explicit geometric information does help the tasks
of classification, part segmentation and semantic segmentation on challenging benchmarks, both
based on clean CAD models of objects, as well as from real world noisy scans of indoor and outdoor
spaces. 

\vspace{-0.8ex}
\subsection{Geometrically Constrained Neighborhood Graph Construction}
\vspace{-0.6ex}

Once the feature description is available, the next step is to define the directed edges between the
points to fully construct the graph. We now detail our second contribution pertaining to the
construction of the graph.

Previous methods construct the graphs using a \knn scheme. Each point is considered, and directed
edges are added from the point to its $k$ nearest neighbors. The graph thus constructed is given as
input to the GNN which performs the final task.

A \knn based graph construction scheme, while being intuitive and easy to implement, has an
inherent problem. It is not robust to sampling frequency difference which can happen often in
practice. Certain object may be densely sampled at some location in the scene, but the same object
when placed at another location in the scene might only be sparsely sampled. Such scenarios can
happen, for instance, in LiDAR based scanning, where the spatial sampling frequency is higher when
the object is nearer to the sensor, but is lower when the object is farther. In such cases, \knn
may not cover the object sufficiently when the sampling density is higher, as all the $k$ nearest
neighbors might be very close together, as shown in Figure~\ref{figIllus} (center), and fall on the
same subregion of the object. Hence, we propose a locality adaptive graph construction algorithm
which promotes coverage in the local graph.

\begin{figure}
\centering
\includegraphics[width=0.4\columnwidth]{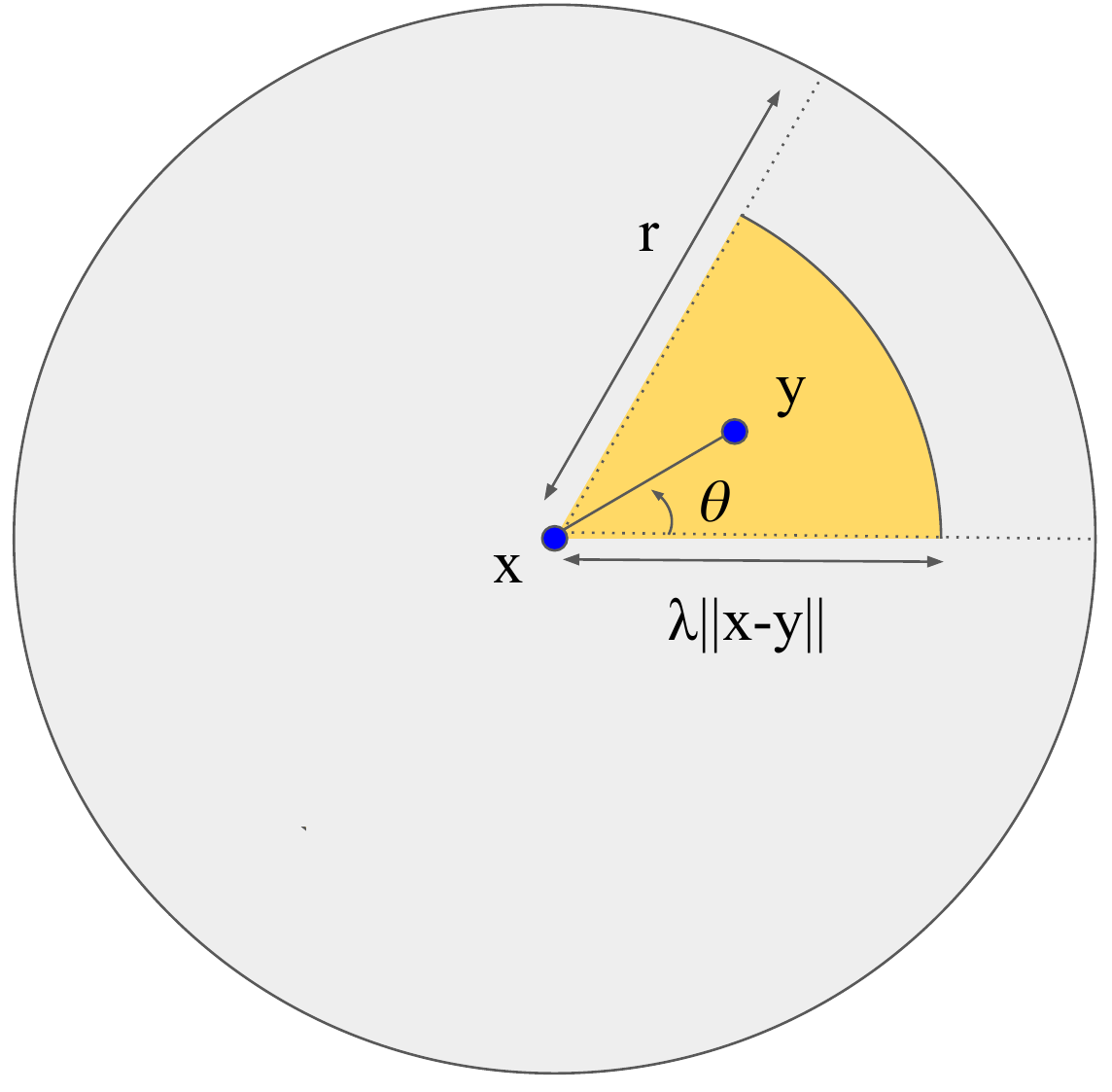}
\vskip -0.1in
\caption{\small{The local neighborhood graph construction constraints in 2D: Once a point $y$ is selected,
for a point $x$, points from the yellow region, \ie within an angle $\theta$ and a distance
$\lambda\| x-y \|$ are discarded. This encourages coverage, and tackles densely sampled parts, in
the local neighborhood graph. Best viewed in color.}}
\vskip -0.25in
\label{figGraphConstruct}
\end{figure}

The design of the graph construction is inspired by log-polar histograms which have been popular for
many engineering applications.
The main element is illustrated in Figure~\ref{figGraphConstruct} in the
case of 2D points. The construction iterates over each point one by one, and considers the sorted 
nearest neighbors of each point. However, instead of selecting the \knn points, it greedily selects
the neighbors as follows. At point \x, as soon as it selects a point \y, it removes all points in
the region within an angle $\theta$ from the line $\y - \x$ with origin at \x, and with distance
less than $\lambda\|\x-\y\|$, with both $\theta, \lambda$ being tunable hyper parameters, shown as
the region marked in yellow in Figure~\ref{figGraphConstruct}. The simple intuition being that once
a point from a local neighborhood has been sampled, we look for points which are sufficiently away
from that local neighborhood. This ensures coverage in the local graph construction and improves
robustness to sensor sampling frequency artefacts.

To illustrate the construction with a small toy example, we refer to Figure~\ref{figBlockDiag}
(step 2). In the usual \knn based graph construction, \wrt point $1$, the points which would have been
selected would be $2,3,4,5,6,7$ (for $k=6$), which would concentrate the local neighborhood graph
only on one region of the space. In contrast, with the proposed method, with appropriate $\theta,
\gamma$, the selection of points $3,4,6$ can be avoided due to the presence of $2,5$, and the final
six selected points would be $2,5,7,8,9,10$ which are well distributed in the space around the main
point. Hence the problem of oversampling in some regions by the sensor can be avoided.

\vspace{-1ex}
\subsection{Use with GNN}
Given the above point descriptions, as well as graph edge constructions, we have a fully specified
graph $\G = (\V,\E)$ derived from the point cloud data. This graph is general enough to be used with
any GNN algorithm, while being specific to 3D point clouds as the vertex descriptions contain local
geometric information, and the edges have been constructed in a geometrically constrained way.
We stress here that while we give results with one specific choice of GNN in this
paper, our graph can be used with any available GNN working on the input graph \G.

\vspace{-0.8ex}
\section{Experimental Results}
\label{secExp}

We now give the results of the proposed methods on challenging benchmarks.

We denote the proposed methods as (i) \emph{ours (representation)}, when we incorporate the
proposed local geometric representation, and (ii) \emph{ours (representation + graph)} when we
incorporate the proposed geometrically constrained graph construction as well. 

\vspace{-0.6ex}
\subsection{Datasets and Experimental Setup}
We evaluate the proposed method on the following datasets ModelNet40~\cite{wu20153d}, ShapeNet~\cite{yi2016scalable}, ScanNet~\cite{dai2017scannet}, Stanford Large-Scale 3D Indoor Spaces (S3DIS) scans~\cite{armeni20163d}, and Paris-Lille-3D (PL3D)~\cite{roynard2018paris}. We follow standard settings for training and evaluation on respective datasets. We use Dynamic Edge Convolutional network (DGCNN) \cite{wang2019dynamic} as the baseline network. Further, the input to the MLP is the 9D vector, it contains two FC layers with $18$ and $9$ units respectively.

\vspace{-0.6ex}
\subsection{Parametric Study}
Table~\ref{tabNN} gives the performances of the methods with different number of nearest neighbors.
We see that for the baseline method, DGCNN with \knn based graph construction, the performance
initially increases with an increase in $k$ but then the drops a little, \eg mean class accuracy
goes from $88.0$ to $90.2$ for $k=5,20$, but then drops to $89.4$ for $k=40$. The authors in DGCNN
\cite{wang2019dynamic} attributed this to the fact that for densities with large $k$ the Euclidean
distance is not able to approximate the geodesic sufficiently well. However, for the proposed method
the performance improves by a modest amount and does not drop, \ie $89.8, 93.1, 93.8\%$ for
$k=5,20,40$ respectively. This indicates that with explicit geometric information provided as input,
the geometry is relatively better maintained \cf DGCNN. As the performance increase from $k=20$ to
$k=40$ is modest for our method, we use $k=20$ as a good tradeoff between performance and time.

The values of $\theta=\frac{\pi}{6}$ and $\lambda=1.25$ were set based on validation experiments on a subset of the training and validation data for all experiments reported. 

\begin{table}[t]
\center
\caption{\small{Results on ModelNet40 classification with varying number of nearest neighbors (\#NN) for graph construction}}
\label{tabNN}
\resizebox{\columnwidth}{!}{
\begin{tabular}{c|cc|cc|cc|cc}
\toprule
\# NN &
\multicolumn{2}{c|}{{DGCNN (\knn)}} &
\multicolumn{2}{c|}{{ours (repr.) + \knn}} &
\multicolumn{2}{c|}{{ours (graph)}} &
\multicolumn{2}{c}{{ours (repr. + graph)}} \\ \hline
$k$ & \multicolumn{1}{c}{m-acc(\%)} & \multicolumn{1}{c|}{ov-acc(\%)} & \multicolumn{1}{c}{m-acc(\%)}
& \multicolumn{1}{c|}{ov-acc(\%)} & \multicolumn{1}{c}{m-acc(\%)} & \multicolumn{1}{c|}{ov-acc(\%)} 
& \multicolumn{1}{c}{m-acc(\%)} & \multicolumn{1}{c}{ov-acc(\%)} \\ \hline
\midrule
5  & 88.0 & 90.5 & 88.9 & 91.1 & 88.3 & 90.8 & 89.8 & 91.9 \\ 
10 & 88.9 & 91.4 & 90.1 & 92.4 & 89.2 & 92.6 & 91.6 & 93.9 \\ 
20 & 90.2 & 92.9 & 92.0 & 95.1 & 90.8 & 94.4 & 93.1 & 95.9 \\ 
40 & 89.4 & 92.4 & 91.4 & 94.6 & 91.2 & 95.1 & 93.8 & 96.6 \\ 
\hline
\end{tabular}
}
\vspace{-2ex}

\end{table}

\begin{table}[h]
\begin{center}
\caption{\small{Comparison with state of the art methods on ModelNet40 classification and ShapeNet part segmentation datasets.}}
\label{tabModelNet40Cls}
\resizebox{\linewidth}{!}{
\begin{tabular}{l|c c c r}
\hline
  Method & \multicolumn{2}{c}{ModelNet40} & ShapeNet \\ \hline
    & m-acc(\%)  & ov-acc(\%) & mIoU(\%) \\
\hline

PointNet \cite{qi2016pointnet}                   & 86.0 & 89.2 & 83.7\\
PointNet++ \cite{qi2017pointnet++}                    & - & 90.7 & 85.1 \\
Kd-net \cite{klokov2017escape}
    & - & 90.6  & 82.3 \\

DensePoint \cite{zhao20193d} & - & 93.2 & 86.4\\
LP-3DCNN \cite{kumawat2019lp} & - & 92.1 & -\\
DGCNN \cite{wang2019dynamic}                         & 90.2 & 92.9 & 85.2 \\
DGCNN (2048 pts) \cite{wang2019dynamic}                & 90.7 & 93.5 & - \\
LDGCNN \cite{zhang2019linked} & 90.3 & 92.9 & 85.1 \\
KPConv-rigid \cite{thomas2019kpconv} & - & 92.9 & 86.2 \\
KPConv-deform \cite{thomas2019kpconv} & - & 92.7 & 86.4 \\
GSN~\cite{xu2020geometry} & - & 92.9 & - \\
GSN (2048 pts)~\cite{xu2020geometry} & - & 93.3 & - \\
LS~\cite{lyu2020learning} & - & - & 88.8 \\
PointASNL~\cite{yan2020pointasnl} & - & 93.2 & - \\
SPH3D-GCN (2048 pts)~~\cite{lei2020spherical} & 88.5 & 91.4 & 86.8 \\
FKAConv~\cite{boulch2020lightconvpoint} & 89.9 & 92.5 & - \\
FKAConv (2048 pts)~\cite{boulch2020lightconvpoint} & 89.7 & 92.5 & 85.7 \\
ConvPoint~\cite{boulch2020convpoint} & 88.5 & 91.8 & 85.8 \\
ConvPoint (2048 pts)~\cite{boulch2020convpoint} & 89.6 & 92.5 & - \\
\hline
Ours (repr.)                        & \textbf{92.0} & \textbf{95.1} & 87.4\\
Ours (repr.\ + graph)                       & \textbf{93.1} & \textbf{95.9} & \textbf{89.1} \\
Ours (repr.\ + graph) - 2048 pts                       & \textbf{94.1} & \textbf{96.9} & - \\
\hline
\end{tabular}
}
\end{center}
\vspace{-2ex}
\end{table}

\vspace{-3ex}
\subsection{Comparison with state of the art}
\vspace{-1.5ex}
\heading{Classification on ModelNet40.} The classification results on the ModelNet40 datasets are shown in Table \ref{tabModelNet40Cls} with comparison against methods based on 3D input. In
the last block the first two rows of our method are with $1024$ points while the third (last) is with $2048$ points.  Our full method achieves the best results, \eg $95.9\%$ overall class accuracy, compared to many recent
competitive methods such as PointASNL ($93.2\%$), RS-CNN ($93.6\%$) and the baseline DGCNN ($92.9\%$). We also see that our method improves more when the point sampling is increased from $1024$ to
$2048$ with mean accuracy going from $93.1\%$ to $94.1\%$, \cf  DGCNN $90.2\%$ to $90.7\%$ for the same settings. 

\heading{ShapeNet Part Segmentation.} Table \ref{tabModelNet40Cls} (Column 4) shows results on ShapeNet part segmentation. Similar to the classification case, we achieve higher results than many recent methods. Our representation only method achieves $87.4\%$ mIoU \cf $85.2\%$ of DGCNN \cite{wang2019dynamic}. While
our full, representation and graph construction based method achieves $89.1\%$ which is higher than recently proposed learning to segment (LS) by $0.3\%$. However, LS projects the 3D points to 2D images and uses U-Net segmentation, while the proposed method directly computes the part labels on 3D points.

\heading{Semantic Segmentation.} Table \ref{tab:semanticseg} show results of semantic segmentation on ScanNet, Stanford Indoor Spaces Dataset (S3DIS) and  Paris-Lille-3D datasets (PL3D). On S3DIS dataset, we outperform all the compared methods \wrt mIoU and overall accuracy (ov-acc) on Area-5 protocol and \wrt overall accuracy on k-fold evaluation protocol. Further, we also outperform other methods \wrt overall accuracy on PL3D dataset. We outperform the next best method on PL3D (KPConv-deform, ConvPoint) by $2.6\%$ on mIoU while lag behind FKAConv by $4.2\%$, however, we outperform FKAConv on S3DIS (k-fold) by $1.7\%$ \wrt mIoU. On ScanNet we achieve an mIoU of $72.4\%$ lagging behind MinkowskiNet ($73.6\%$) and Virtual MVFusion ($74.6\%$), however, we outperform these methods on S3DIS by $\sim4\%$.  

Overall, we observe that our method consistently provides results amongst top-$3$ methods on all the compared benchmarks, demonstrating the robustness of the proposed modifications to datasets with significant noise \eg ScanNet and well defined CAD models \eg ModelNet40.

\begin{table}[!t]
\caption{\small{Comparison with state of the art methods for 3D scene segmentation on Scannet v2, Stanford 3D Indoor Spaces (S3DIS) and Paris-Lille-3D (PL3D) datasets. 
}}
\vspace{-2ex}
\label{tab:semanticseg}
\resizebox{\columnwidth}{!}{
\begin{tabular}{ l|c c c c c c c }
\hline
Method & Scannet & \multicolumn{2}{c}{S3DIS (k-fold)} & \multicolumn{2}{c}{S3DIS (Area 5)} & \multicolumn{2}{c}{PL3D}\\
\hline
& mIoU & mIoU & ov-acc & mIoU & ov-acc & mIoU & ov-acc \\
\hline
PointNet \cite{qi2016pointnet} & - & 47.6 & 78.6 & 41.1 & - & 40.2 & 93.9 \\
PointNet++ \cite{qi2017pointnet++} & 33.9 & 54.5 & 81.0 & - & - & 36.1 & 88.7 \\
PointCNN \cite{li2018pointcnn} & 45.8 & 65.4 & 88.1 & 57.3 & 85.9 & 65.4 & -\\
KPConv-rigid~\cite{thomas2019kpconv} & 68.6 & 69.6 & - & 65.4 & - & 72.3 & -\\
KPConv-deform~\cite{thomas2019kpconv} & 68.4 & \textbf{70.6} & - & 67.1 & - & 75.9 & -\\
DGCNN~\cite{wang2019dynamic} & - & 56.1 & 84.1 & - & - & 62.5 & 97.0\\
MinkowskiNet~\cite{choy20194d} & 73.6 & - & - & 65.4 & - & - & -\\
HDGCNN~\cite{liang2019hierarchical} & - & 66.9 & - & 59.3 & - & 68.3 & - \\
DPC \cite{engelmann2019dilated} & 59.2 & - & - & 61.3 & 86.8 & - & - \\
Virtual MVFusion \cite{kundu2020virtual} & \textbf{74.6} & - & - & 65.3 & - & - & -\\
JSENet~\cite{hu2020jsenet} & 69.9 & - & - & 67.7 & - & - & -\\
FusionNet~\cite{zhang12356deep} & 68.8 & - & - & 65.38 & - & - & -\\
DCMNet~\cite{schult2020dualconvmesh} & 65.8 & 69.7 & - & 64.0 & - & - & -\\
PointASNL~\cite{yan2020pointasnl} & 63.0 & 68.7 & - & - & - & - & - \\
SPH3D-GCN~\cite{lei2020spherical} & 61.0 & 68.9 & 88.6 & 59.5 & 87.7 & - & - \\
SegGCN~\cite{lei2020seggcn} & 58.9 & 68.5 & 87.8 & 63.6 & 88.2 & - & - \\
ResGCN-28~\cite{li2019deepgcns} & - & 60.0 & 85.9 & - & - & - & - \\
FKAConv~\cite{boulch2020lightconvpoint} & - & 68.4 & - & - & - & \textbf{82.7} & - \\
ConvPoint~\cite{boulch2020convpoint} & - & 69.2 & 88.8 & - & - & 75.9 & - \\

\hline
Ours (repr.+graph)  & 72.4 & 70.1 & \textbf{89.8} & \textbf{69.4} & \textbf{89.4} & 78.5 & \textbf{98.0}\\
\hline
RANK & \textbf{\textcolor{cyan}{3}} & \textbf{\textcolor{blue}{2}} & \textbf{\textcolor{red}{1}} & \textbf{\textcolor{red}{1}} & \textbf{\textcolor{red}{1}} & \textbf{\textcolor{blue}{2}} & \textbf{\textcolor{red}{1}}\\
\hline
\end{tabular}
}
\vspace{-2ex}

\end{table}

\vspace{-1ex}
\subsection{Qualitative Results}
\begin{figure}[t]
\includegraphics[width=\columnwidth]{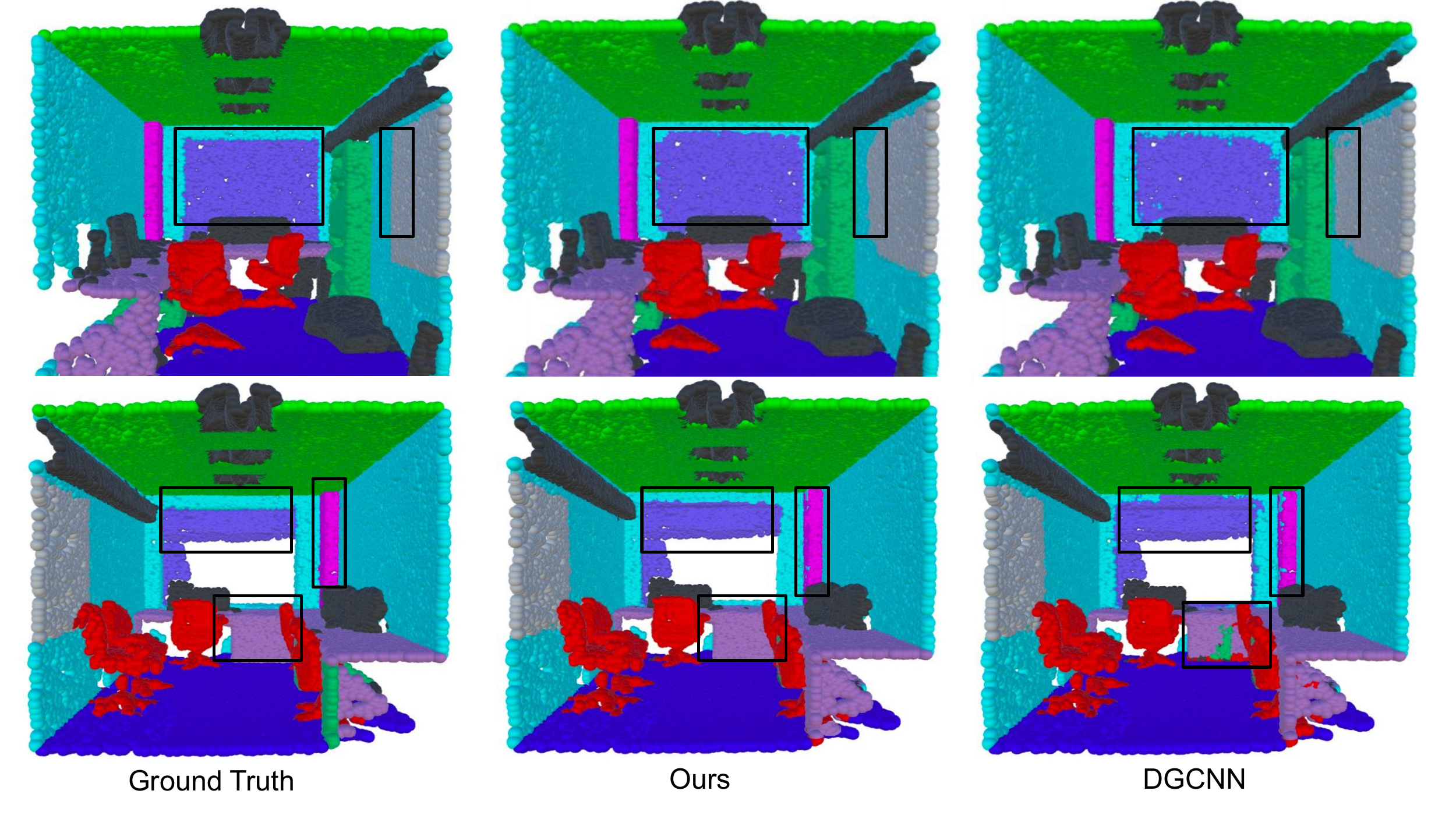}
\vskip -0.1in
\caption{Qualitative results on S3DIS dataset, notice the fine structures captured by our method and
missed by DGCNN, in the annotated black boxes. Best viewed zoomed in color.}
\label{figQuals3dis}
\vskip -0.2in
\end{figure}

Figure \ref{figQuals3dis} shows semantic segmentation results on S3DIS. In general, we observe that
the proposed technique provides closer to ground-truth segmentation especially at the boundaries. In
first row, we observe that the structures on the wall are better constructed by the proposed method 
as compared to DGCNN especially around the edges. In the box at center, we observe that the proposed
method marks the points on the wall behind the table correctly while DGCNN incorrectly
labels them as wall. Similarly, in second row, the labels in the box at top are closer to the ground
truth with a clear separation from the wall. In the bottom box, we observe that the points
on the table are clearly labelled by our method while DGCNN labels many points as walls. These
results indicate that the proposed method is robust in narrow regions and on fine structures due to
the explicit introduction of geometrical features to GCNN. 

\begin{table}[h]
\caption{\small{Ablation studies on impact of various components of the proposed method.}}
\vspace{-2ex}
    \label{tabAblAll}
\centering
   
    \resizebox{\linewidth}{!}{
    \begin{tabular}{ccc|cccccc}
        \hline
        & & & \multicolumn{2}{c}{  ModelNet40   } &   ShapeNet   &
        \multicolumn{2}{c}{  S3DIS   }  \\ \hline
        $\alpha, \beta$ & \; \; $\gamma$ \; \; & graph \quad \; &   m-acc   & ov-acc &
        mIoU &  mIoU  & ov-acc  \\
        \hline
        $\times$ & $\times$ & $\times$          & 90.2 & 92.9  & 85.2 & 56.1 & 84.1\\
        \checkmark & $\times$ & $\times$        & 90.8 & 93.4  & 86.1 & 57.8 & 84.7\\
         $\times$ & \checkmark &$\times$        & 90.7 & 93.8  & 85.9 & 57.8 & 85.1\\
         $\times$ & $\times$  & \checkmark      & 90.8 & 94.4  & 86.2 & 59.4 & 86.6\\
         \checkmark & $\times$  & \checkmark    & 92.1 & 95.1  & 87.1 & 61.2 & 87.4\\
         $\times$   & \checkmark & \checkmark   & 91.8 & 94.9  & 86.8 & 62.1 & 87.4\\
        \checkmark   & \checkmark & $\times$    & 92.0 & 94.0  & 86.6 & 65.5 & 88.1 \\
        \checkmark   & \checkmark & \checkmark  & 93.1 & 95.9  & 89.1 & 69.4 & 89.4 \\
        \hline
    \end{tabular}
    }
    \vspace{-2ex}
     
\end{table}

 \vspace{-1.5ex}
\subsection{Ablation Experiments}\label{subsec:ablation}
 \vspace{-0.5ex}

We perform ablation experiments to give the complete results with and without including the different
components of the proposed method.

For ablation on absence of various local features, we only set the corresponding elements of the
input vector to the MLP and train the projection MLP as usual. 

\heading{Ablation on Classification.}
In Table \ref{tabAblAll} columns 4 and 5, starting from a baseline of $92.9\%$ overall accuracy, we see that the local
geometric features help individually, but their combination is better at $94.9\%$. Similarly, the
graph construction method gives $94.4\%$ alone but improves with each of the geometric features
added, and performance is best when all the features as well as graph construction are used
together, at $95.9\%$. 

\heading{Ablation on Part Segmentation.}
In Table \ref{tabAblAll} column 6, we see similar trends as in the classification case, where each of the geometric features
contribute ($86.1\%, 85.9\%$), the graph construction helps when used alone ($86.2\%$), and the
full combination achieves the best performance at $89.1\%$ mIoU, \cf the baseline at $85.2\%$ mIoU.

\heading{Ablation on Semantic Segmentation.} In Table \ref{tabAblAll} columns 7 and 8, we observe that
with the graph construction only the mIoU increases from $56.1\%$ to $59.4\%$ while it increases to
$65.5\%$ with only the combination of local features. However, the full combination provides the
best performance at mIoU of $69.4\%$ and overall accuracy of $89.4\%$.

\begin{table}[h]

\begin{center}
\caption{\small{Ablation study on the training convergence of models}}
\label{tabEpochs}
\resizebox{\linewidth}{!}{
    \begin{tabular}{cc|cccccc}
    \hline
    & & \multicolumn{2}{c}{\; ModelNet40 \; } & \multicolumn{2}{c}{\; ShapeNet \; }  & \multicolumn{2}{c}{\; S3DIS \; } \\
    \hline
     $\alpha, \beta, \gamma$  & \; \; graph \; \; & \quad tr.\ epochs \quad & frac.& \quad tr.\
     epochs \quad & frac.& \quad tr.\ epochs \quad & frac.\\
     \hline
    $\times$ & $\times$         & 233  & 1.00 & 190 & 1.00 & 258 & 1.00 \\
    \checkmark & $\times$       & 205  & 0.87 & 174 & 0.91 & 243 & 0.94 \\
    $\times$ & \checkmark       & 164  & 0.70 & 151 & 0.79 & 215 & 0.83 \\
     \checkmark & \checkmark    & 139  & 0.60 & 133 & 0.70 & 191 & 0.73 \\
     \hline
    \end{tabular}
}
\vspace{-2em}
\end{center}
\end{table}

\heading{Ablation on Model Convergence} 
The number of epochs required for the different models to converge are shown in Table
\ref{tabEpochs}. We observe that using the proposed geometric representation results in reduction by
$13\%$, $9\%$ and $6\%$ epochs for classification, part segmentation and semantic segmentation
respectively. With the proposed geometrically aware graph construction, we observe a reduction of
$30\%$ for classification, $21\%$ for part segmentation and $17\%$ for semantic segmentation. Hence,
providing a geometrically refined local connectivity information to the graph results in significant
reduction in the number of epochs required by the network. Finally, the combination of the proposed
geometric representation and the proposed graph construction results in a reduction of $40\%, 30\%,
27\%$ epochs required for the tasks respectively.

The actual training time required to train our full method on ModelNet40 is $3.2$ hours as
compared to $5$ hours required by DGCNN, providing a training speed-up of nearly $36\%$, which is
similar to the epoch reduction of $40\%$, \ie less epochs translate directly into reduction
in training time. 

\heading{Ablation on MLP.} With only 3D coordinates as input to the MLP, the m-acc on ModelNet40 is $90.9\%$ \cf $90.2\%$ (DGCNN). With 9D vector as a direct input without MLP to DGCNN, the m-acc is $91.1\%$ \cf $92.0\%$ with MLP (Table \ref{tabModelNet40Cls}, Ours(repr.)). This shows that MLP is able to extract a robust geometric representation in the feature space.

\vspace{-1ex}
\section{Conclusion}
\label{secConclusion}
We proposed two simple but important contributions, to be used within the framework of graph neural
networks (GNN) for point cloud processing, on (i) representation of points using local geometric
properties, and (ii) construction of the neighborhood graph using geometric constraints to improve
coverage. We also reported highly competitive results, better than many recent
state-of-the-art methods on both synthetic and real-world datsets showing robustness of the proposed approach to noisy data. Our work highlights the fact that the current generation of GNN based methods, for
3D point cloud processing, are not able to fully capture the local geometric information, and hence
benefit from having that as input explicitly. Further, recently Liu \etal~\cite{liu2020closer} showed that, the local aggregation operators used in various point cloud processing techniques, if carefully tuned, provide similar performances. Parida \etal~\cite{parida2021beyond} showed that using region/point based properties from echoes or type of material can help in learning more robust representations. Therefore, the proposed modifications, with explicit prior on geometry, can be extended to other point cloud based deep networks and potentially motivate future works with simpler and more efficient networks for processing point clouds.

\bibliographystyle{IEEEtran}
\bibliography{IEEEfull}

\end{document}